\pdfoutput=1

\documentclass[11pt]{article}
\usepackage[table]{xcolor}

\usepackage{acl}

\usepackage{times}
\usepackage{latexsym}
\usepackage{amssymb}
\usepackage[T1]{fontenc}
\usepackage{comment}
\usepackage[utf8]{inputenc}

\usepackage{microtype}

\usepackage{graphicx}
\usepackage{tabularx}
\usepackage{multirow}
\usepackage{amsmath}
\usepackage{diagbox}
\usepackage{enumitem}

\title{Memory-efficient NLLB-200: Language-specific Expert Pruning of a Massively Multilingual Machine Translation Model}

\author{Yeskendir Koishekenov$^{*1,2}$ Alexandre Berard$^{1}$ Vassilina Nikoulina$^{1}$ \\\
        $^{1}$NAVER LABS Europe \\  $^{2}$University of Amsterdam\\
        \texttt{\{first.last\}@naverlabs.com} \\
        \texttt{yeskendir.koishekenov@student.uva.nl}
        }

\usepackage{todonotes}
\begin{document}
\maketitle
\begingroup\def\thefootnote{*}\footnotetext{Work done during an internship at NAVER LABS Europe}\endgroup

\begin{abstract}



The recently released NLLB-200 is a set of multilingual Neural Machine Translation models that cover 202 languages. The largest model is based on a Mixture of Experts architecture and achieves SoTA results across many language pairs. It contains 54.5B parameters and requires at least four 32GB GPUs just for inference.
In this work, we propose a pruning method that enables the removal of up to 80\% of experts without further finetuning and with a negligible loss in translation quality, which makes it feasible to run the model on a single 32GB GPU. Further analysis suggests that our pruning metrics can identify language-specific experts. 

\end{abstract}

\section{Introduction}
The Transformer \cite{vaswani2017attention} has become the dominant modeling paradigm in Natural Language Processing tasks.
Many subsequent advances in the field came from increasing the computational budget, training data, and model size.
Neural Machine Translation was not an exception, where massively multilingual NMT \cite{aharoni2019massively, fan2021beyond, tang2020multilingual, zhang2020improving} demonstrated promising results, while attempting to overcome the curse of multilinguality \cite{conneau2019unsupervised} by scaling up model size. 

However, increasing the parameter size exacerbates the cost of training \cite{yang2019xlnet, strubell2019energy, patterson2021carbon} and hurts the memory footprint and inference latency \cite{dai2019transformer, fan2021beyond, wang2022deepnet}. 
Sparsely-gated Mixture-of-Experts (MoE) models are an efficient alternative to dense models \cite{lepikhin2020gshard,fedus2021switch,riquelme2021scaling}. For example, \citet{glam2021} demonstrates that an MoE language model results in a 7x larger model compared to GPT-3, but requires only 30\% of its energy for training and half of its FLOPs at inference.

Mixture-of-Experts models are neural networks whose set of parameters is partitioned into experts. Contrary to dense models, where all network parameters are used for every input, an MoE model activates different parts of the network, the experts, depending on the input, which is typically done by a gating mechanism at the token level. MoE models are computationally efficient due to expert parallelism \cite{fedus2021switch} across a large number of GPUs, by having each GPU hold a subset of all experts and communicate with the other GPUs when it needs expert outputs for its local batch.

In NLLB-200\footnote{In what follows, NLLB-200 refers to the 54.5B-parameter MoE NLLB model, unless specified otherwise.}\cite{costa2022no}, a load balancing regularizer in the objective function \cite{shazeer2017outrageously} promotes equal distribution of the tokens across experts. This encourages the model to use all the experts and ensures that all GPUs are used equally for the sake of computational efficiency. However, considering a large number of experts, it does not guarantee that all experts will be equally activated for a particular pair of languages at inference. It raises a research question: \textit{are there language-specific experts in multilingual MoE models?} If this is the case, we may be able to prune such models without loss of translation quality for the language pairs of our interest. Reducing memory usage would be useful for a model like NLLB-200, which normally requires at least four 32GB GPUs at inference.

In this work, we define metrics to assess the importance of each expert and prune the least important experts at inference. We aim to avoid fine-tuning because of its computational cost. In an ideal scenario, we would like to be able to identify the important experts in an MoE model so that practitioners can deploy large models, such as NLLB-200, on a single GPU.  We summarize our main contributions as follows:
\begin{itemize}[noitemsep, nolistsep]
    \item We propose a pruning strategy that can remove 80\% of experts in the NLLB-200 model without further finetuning and with a negligible loss in translation quality;
    \item We find that the decoder experts can be pruned more aggressively than the encoder experts;
    \item We show the emergence of language-specific experts in the NLLB-200 model;
    \item We demonstrate that the important language-specific experts in the decoder are shared between linguistically related languages;
    \item We release the ids of the pruned experts, along with other experts' gathered statistics so that anyone with a single 32GB GPU can use NLLB-200 at inference.\footnote{\url{https://europe.naverlabs.com/research/natural-language-processing/nllb-200-expert-pruning}}

\end{itemize}
\section{Related work}

The concept of Mixture-of-Experts models in machine learning dates back to the works of \citet{jacobs1991adaptive, jordan1994hierarchical}. Most recent versions were inspired by \citet{shazeer2017outrageously}, who achieved state-of-the-art language modeling and translation results with the largest model at that time. Combined with the Transformer model, MoE models grew in popularity \cite{lepikhin2020gshard,fedus2021switch}. Beyond natural language processing, MoE models showed a large success in computer vision \cite{puigcerver2020scalable}, speech recognition \cite{you2021speechmoe}, multi-modal learning \cite{mustafa2022multimodal}, and diffusion models \cite{feng2022ernie, balaji2022ediffi} to name a few. For a more detailed survey of MoE models, we refer readers to \citet{yuksel2012twenty} and \citet{fedus2022review}. 

Despite the recent successes, large MoE models require a lot of memory and the contribution (or roles) of experts is under-explored. \citet{chen2022task} showed that the contributions of experts of a pre-trained MoE model in different tasks such as MNLI, CoLA, and SQuAD are quite different. Moreover, they converted a large sparse MoE model pre-trained on a general task to a single-expert dense model by fine-tuning the most `professional' expert and dropping the other experts. It demonstrates that experts do not contribute equally to the performance and some are more important than others. \citet{zoph2202st} also studied different expert specializations such as sentinel tokens, punctuation, conjunctions and articles, and even languages.  They concluded that experts in the encoder exhibit specialization, in contrast to the decoder, but not by language. According to the authors, their mechanism of token routing and load balancing prevents language specialization. 


\citet{kudugunta2021beyond} train study routing mechanisms at different levels of granularity and show that task-level experts (i.e., per language) can achieve similar performance as token-level experts. However, this work assumes that the model is trained this way, while our own work attempts to prune an existing token-level MoE model at inference without re-training it.

There have been a number of attempts to compress existing massively multilingual NMT models \cite{costa2022no,Mohammadshahi2022WhatDC, Mohammadshahi2022SMaLL100IS}. However, to the best of our knowledge, none of them explicitly studied expert pruning and the emergence of language-specific experts in a large MoE model like we do. There has been a related line of works on pruning attention heads in transformer models \cite{Michel19, voita2019analyzing}, demonstrating linguistically-interpretable roles of attention heads \cite{voita2019analyzing, jo2020roles} and the emergence of language-specific attention heads \cite{kim2021multilingual, held2022shapley}. Understanding the role of attention heads helps carefully remove the least important ones without damage to translation quality.

Closest to our work, \citet{kim2021scalable} tried to prune a machine translation MoE model by keeping the most activated experts,\footnote{Equivalent to our \textit{activity} pruning metric.} but did not manage to preserve performance without further fine-tuning.

Even though it has been shown that multilingual NMT models benefit from a larger number of experts \cite{costa2022no}, to the best of our knowledge, our work is the first to study whether any language-specific experts emerge in a massively multilingual Mixture-of-Expert model for NMT, and how can redundant (or non-relevant) experts be pruned. 

\section{Background}
\label{prelim}

\subsection{Mixture-of-Experts models}

Sparsely-gated Mixture-of-Experts (MoE) models activate a subset of their parameters per input token, contrary to dense models, where the entire network is used for each input token. Therefore, the total amount of parameters can be significantly increased because the computation cost per token becomes only proportional to the size of the activated sub-network, not the total model size. An increased number of parameters unlocks significant representational capacity. Allocating different devices for different experts and running them in parallel (i.e., expert parallelism, \citealp{fedus2021switch}), in combination with data parallelism makes MoE computationally efficient and highly scalable \cite{fedus2021switch, lepikhin2020gshard}.

In the MoE Transformer models proposed by \citet{lepikhin2020gshard}, the FFN sublayers in the dense model are replaced with MoE layers. An MoE layer takes an input token representation $x_t$ and then routes it to the top-$k$ experts selected from a set $\{E_{i}\}^{N}_{i=1}$ of $N$ experts thanks to a gating network:
\begin{equation}
    G_t = softmax(W_g \cdot x_t)
\end{equation}
Where $W_g \in \mathbb{R}^{N\times{}d}$ is a learned parameter.
The output of the MoE layer is a weighted sum of the outputs of the $k$ selected experts $\mathcal{E}$:
\begin{equation}
    y_t = \frac{1}{\sum_{i \in \mathcal{E}} G_{t,i}} \sum_{i \in \mathcal{E}} G_{t,i}E_{i}(x_t)
\end{equation}

\subsection{NLLB-200}

\textit{No Language Left Behind} (NLLB-200) is a set of massively multilingual NMT models that can translate to and from 202 languages \cite{costa2022no}, including many very low resources languages. Models of varying sizes have been released. The largest one is a Mixture-of-Experts model and has 54.5B parameters. A dense model of 3.3B models is also available, which has the same architecture as the 54.5B MoE model without the experts. In this work, we will attempt to prune the experts from the 54.5B model while using the 3.3B variant as a \textit{lower-bound} baseline.\footnote{If the pruned models' performance is worse than the 3.3B baseline, there is no point in using the MoE model, which is larger and more cumbersome to use.}

In the 54.5B MoE model, every 4\textsuperscript{th} FFN sublayer -- in both the encoder and decoder -- is replaced by an MoE layer, starting at the 4\textsuperscript{th} layer (this makes 12 layers with experts). Each MoE layer consists of 128 experts (1536 experts in total) with the same architecture as an FFN sublayer, and has its own gating network, following the top-$k$ gating algorithm of \citet{lepikhin2020gshard} and selecting the top-$2$ experts per token without any randomization. The model was trained with a linear combination of label-smoothed cross-entropy \cite{szegedy2016rethinking} with an auxiliary load balancing loss \cite{shazeer2017outrageously}, which encourages tokens to be uniformly distributed across experts.

\paragraph{Memory usage.}

The 3.3B and 54.5B models are Transformers with an embedding dimension of 2048, an FFN dimension of 8192, 16 attention heads, 24 encoder layers, and 24 decoder layers.
When storing their parameters in half precision, the 3.3B dense model and 54.5B MoE model take respectively 6.2GiB and 101.5GiB of memory. Each expert has 33.6M parameters, representing 51.6B parameters in total or 96GiB of memory. While the 3.3B model can easily run on a single GPU, the 54.5B model requires at the very least 4 32GB GPUs to run. To maximize efficiency, decoding with the MoE model has to be done with expert parallelism \cite{fedus2021switch}, with each GPU holding a full copy of the ``dense'' parameters (2.9B or 5.5GiB) and $1/N$\textsuperscript{th} of the experts per layer, where $N$ is the number of GPUs.\footnote{This brings the memory usage to 118GiB (or 29.5GiB per GPU) when decoding on 4 GPUs.}
Because of the memory usage of beam search decoding and memory fragmentation, batched decoding actually requires more GPUs in practice (e.g., 6 or 8), or to offload the encoder and decoder to the CPU when they are not used.\footnote{Memory usage can be divided by almost two by encoding the full test set with the encoder and then moving the encoder to CPU and decoder to GPU.}

\section{Our Approach}
We experiment with different experts' pruning metrics and strategies that allow us to select the most relevant experts per language or language pair, and thus significantly reduce the memory usage at inference time of NLLB-200.

\subsection{Expert pruning metrics}
\label{sec:metric}
The pruning metric should quantify the contribution of a given expert to the translation. Intuitively, experts that were more involved in translation should be considered more important.

\textbf{Activity.} We define the \textit{Top 1 activity}, $top_1(e)$, of an expert $e$ as the fraction of tokens routed to this expert as the first choice (i.e., the frequency at which this expert was ranked first by the gating mechanism). 
We also consider the \textit{Top 2 activity} variant, $top_2(e)$, with the fraction of tokens routed to this expert as their first or second choice.

Using only \textit{activity} as an importance metric can be sub-optimal as it does not take into account the gating value assigned to this expert by the model. 

\textbf{Load Balancing.} We experiment with the \textit{load balancing} pruning metric, similar to the load balancing loss used by \citet{costa2022no} to train the MoE model. It is defined as the product of the \textit{activity} and the average gate value: $LB(e) = top_1(e) \times mean(e)$.  

\textbf{Importance.} Following the definition of attention head confidence by \citet{voita2019analyzing}, we define the \textit{confidence} of an expert, $conf(e)$, as its average gate value when it is \textit{ranked first}.
Then, we can define the ``vanilla'' \textit{importance} of an expert as the product of its' \textit{activity }and \textit{confidence}.\footnote{Using confidence alone as a pruning metric has demonstrated very poor performance in our preliminary experiments, and therefore was not retained for the follow up study.} 
\begin{equation}
    imp_{vanilla}(e) = top_1(e) \times conf(e)
\end{equation}
We define \textit{importance} as an improved version of \textit{vanilla importance} with an exponential to smooth the confidence values:
\begin{equation}
    imp(e) = top_1(e) \times \exp{(conf(e))}
\end{equation}

\subsection{Expert statistics granularity}
\label{sec:granularity}
To compute the pruning metrics defined above, for each expert $e\in\{1,\dots,1536\}$\footnote{12 layers with 128 experts each = 1536 experts} we collect the gate statistics, $top_1(e)$, $top_2(e)$,  $mean(e)$ and $conf(e)$, by decoding the validation sets for all language directions.\footnote{We always use beam search with a beam size of 4.} However, these statistics can be aggregated at different granularity levels. Depending on how these statistics are aggregated, we hope to see language-specific experts emerge.
In our experiments, we consider three different granularities:
\begin{itemize}[noitemsep,nolistsep]
    \item \textit{global}: we aggregate the statistics across all language pairs to keep the overall best experts;
    \item \textit{language-pair}: we collect gate statistics for each language pair and thus keep a (potentially) different set of experts for each language pair;
    \item \textit{language-specific}: we aggregate encoder-side statistics per source language and decoder-side statistics per target language, which will let us keep a single set of encoder/decoder experts per source/target language. 
\end{itemize}

\subsection{Expert pruning algorithm}
\label{sec:mechanism}

Using the pruning metrics defined in Section~\ref{sec:metric}, there are different expert pruning strategies that we can adopt. The pruning metric values are normalized to sum to one in each layer, and experts are sorted from most important to least important.

\paragraph{Fixed per layer.} First, the simplest way is to retain a fixed amount of top experts in each layer. For example, 75\% pruning retains 384 out of 1536 experts, which corresponds to 32 experts per layer. In the \emph{balanced} setting, the number of experts per layer is the same in the encoder and decoder (e.g., 32 per layer). In the \textit{unbalanced} setting, we keep a different number of experts in the encoder and decoder (e.g., 40 per encoder layer and 24 per decoder layer).
\paragraph{Global threshold.}
The pruning metrics we defined let us easily prune experts per layer, but not globally. To select \textit{globally best} experts (with no \textit{a priori} on the number of experts per layer) we search for a global threshold $\theta$ such that:

\begin{equation}
    \sum_{k=1}^{12} min(
        n_k \mid 
        \sum_{i=1}^{n_k}{\phi(e^k_i)} \geq{} \theta
    ) = count
\end{equation}

Where $\phi$ is the pruning metric; $k$ the layer id (out of 12 layers with experts); $e^k_i$ the $i$\textsuperscript{th} expert in the sorted list of experts for that layer; and $count$ the desired total number of experts to retain (e.g., 384 for 75\% pruning).
Experts $\{e^k_i\}_{i=1}^{n_k}$ are then retained and the rest are pruned.\footnote{We iterate from 0 to 1 by increments of $0.001$ until we find a value of $\theta$ which satisfies this equation.} In our experiments, we make sure to keep at least 4 experts per layer.\footnote{To be able to decode on up to 4 GPUs and to limit the risk of degenerate behavior because some layers have too few experts. Also since NLLB-200 uses top-2 gating, we need at least 2 experts per layer.}

Our intuition behind this pruning method is to define a constant probability mass (or ``importance'' mass) each layer should have. Keeping only a couple of experts in a layer is fine if they are collectively used a majority of the time. Conversely, some layers may need more experts if expert usage is more uniformly distributed.

\begin{figure}[!t]
    \centering
    \includegraphics[width=\columnwidth]{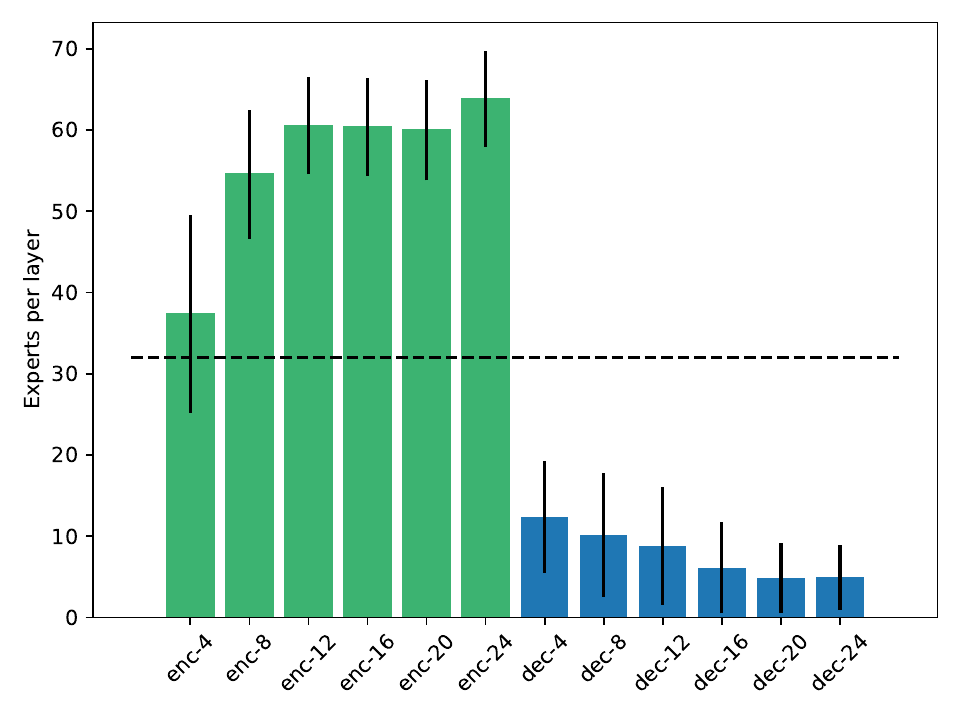}
    \caption{Average number of experts per layer after pruning 75\% of experts with the global threshold algorithm (average activity threshold: 0.69). Pruning is done per language direction and the values are averaged over the 870 directions of the valid set.}  
    \label{fig:experts_per_layer}
\end{figure}

Figure~\ref{fig:experts_per_layer} illustrates how experts are distributed among layers with this approach at 75\% pruning and with the $top_1$ metric. We see that the decoder requires much fewer experts per layer than the encoder to reach the same activity threshold.

We also experiment with a variant of this method, which we call \textbf{Enc/Dec thresholds}, with a fixed amount in the encoder and decoder (e.g., 192 and 192) and thresholds that are defined independently in the encoder and decoder.

\begin{table*}[t]
\begin{center}
\resizebox{\textwidth}{!}{
\begin{tabular}{ c | c | c | c | c | c | c | c | c | c | c | c  } 
Method & Metric & High$\rightarrow$High &High$\rightarrow$Low & High$\rightarrow$V. low & Low$\rightarrow$High & Low$\rightarrow$Low & Low$\rightarrow$V. low & V. low$\rightarrow$High & V. low$\rightarrow$Low & V. low$\rightarrow$V. low & Average \\
\hline
\multicolumn{2}{c|}{3.3B dense model \cite{costa2022no}} & 44.54 & 38.20 & 30.08 & 40.49 & 35.19 & 27.61 & 35.27 & 30.68 & 24.75 & 34.06 \\
\multicolumn{2}{c|}{54.5B MoE model \cite{costa2022no}} & 45.90 & 39.19 & 30.24 & \textbf{42.29} & \textbf{36.35} & 28.18 & \textbf{36.55} & \textbf{32.16} & 24.93 & 35.07 \\
\hline
\multirow{5}{*}{Fixed per layer (balanced)} & Top 1 & 45.52 & 38.75 & 30.13 & 41.51 & 35.50 & 27.92 & 36.09 & 31.68 & 24.90 & 34.64 \\ 
& Top 2 & 44.38 & 37.92 & 29.60 & 40.56 & 34.86 & 27.48 & 35.24 & 30.97 & 24.54 & 33.93 \\ 
& Load balancing & 44.48 & 38.06 & 29.64 & 40.67 & 34.95 & 27.56 & 35.29 & 31.04 & 24.59 & 34.01 \\ 
& Importance (vanilla) & 42.87 & 34.73 & 28.40 & 40.92 & 34.17 & 27.46 & 34.96 & 29.71 & 23.99 & 33.00 \\ 
& Importance & 45.59 & 38.76 & 30.18 & 41.50 & 35.41 & 27.87 & 36.15 & 31.69 & 24.96 & 34.66 \\ 
\hline
\multirow{2}{*}{Global threshold} & Top 1 & 46.01 & 39.28 & 30.44 & 41.91 & 36.18 & 28.21 & 36.40 & 31.97 & 25.06 & 35.03 \\ 
& Importance & \textbf{46.10} & \textbf{39.31} & \textbf{30.46} & 41.99 & 36.25 & \textbf{28.29} & 36.47 & 32.09 & \textbf{25.10} & \textbf{35.09} \\ 
\hline
Fixed per layer (unbalanced) & \multirow{3}{*}{Importance} & 45.79 & 39.00 & 30.33 & 41.80 & 35.76 & 28.12 & 36.36 & 31.93 & \textbf{25.10} & 34.89 \\ 
Enc/Dec thresholds (balanced) & & 45.57 & 38.73 & 30.07 & 41.52 & 35.36 & 27.81 & 36.13 & 31.62 & 24.88 & 34.61 \\ 
Enc/Dec thresholds (unbalanced) & & 45.88 & 38.97 & 30.28 & 41.92 & 35.85 & 28.10 & 36.39 & 31.84 & 25.06 & 34.90 \\ 
\hline
\end{tabular} 
}
\caption{chrF++ valid scores on 30 languages of different pruning algorithms and metrics, with 75\% pruning (i.e., 384 experts are kept in total). The unbalanced approaches keep 240 encoder experts and 144 decoder experts.} 
\label{tab:pruning_algo_and_metric}
\end{center}
\end{table*}

\begin{figure*}
    \centering
    \includegraphics[width=0.4\textwidth]{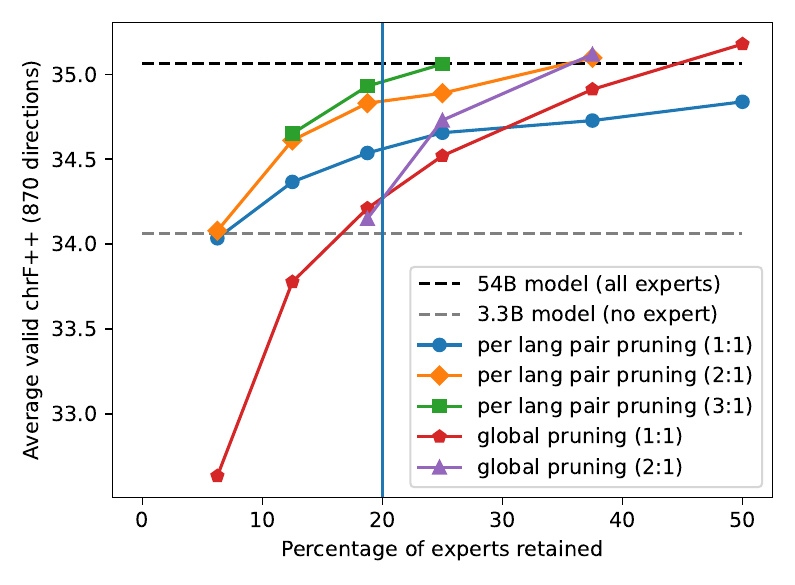}
    \includegraphics[width=0.4\textwidth]{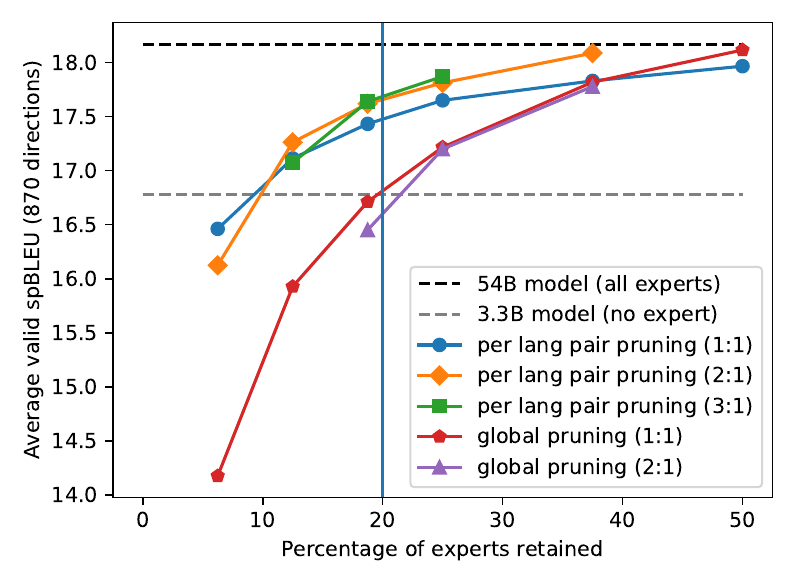}
    \includegraphics[width=0.325\textwidth]{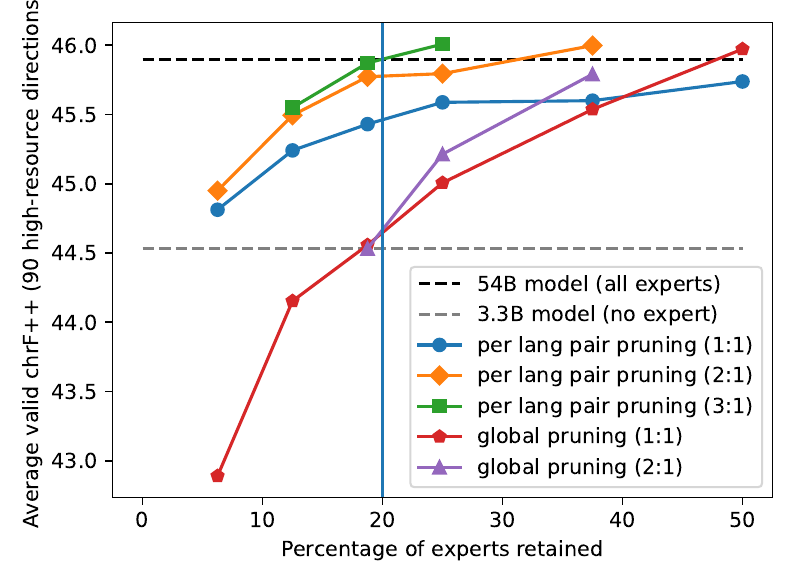}
    \includegraphics[width=0.325\textwidth]{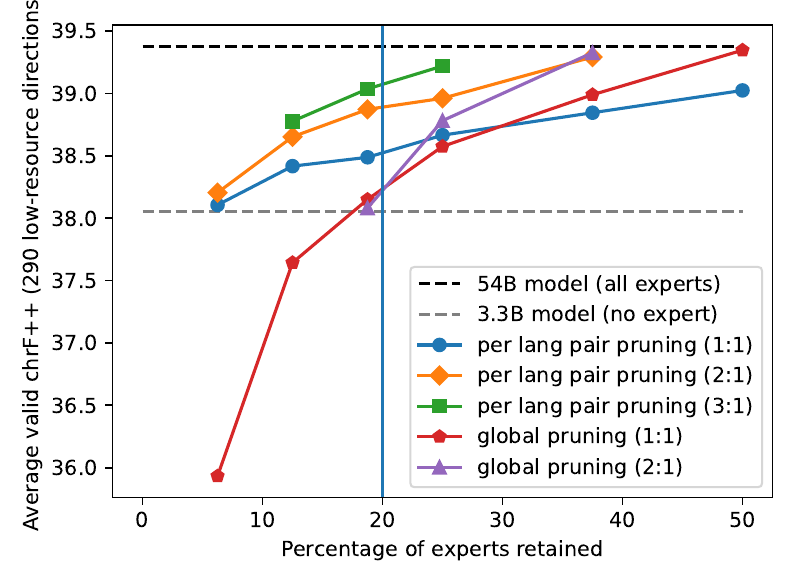}
    \includegraphics[width=0.325\textwidth]{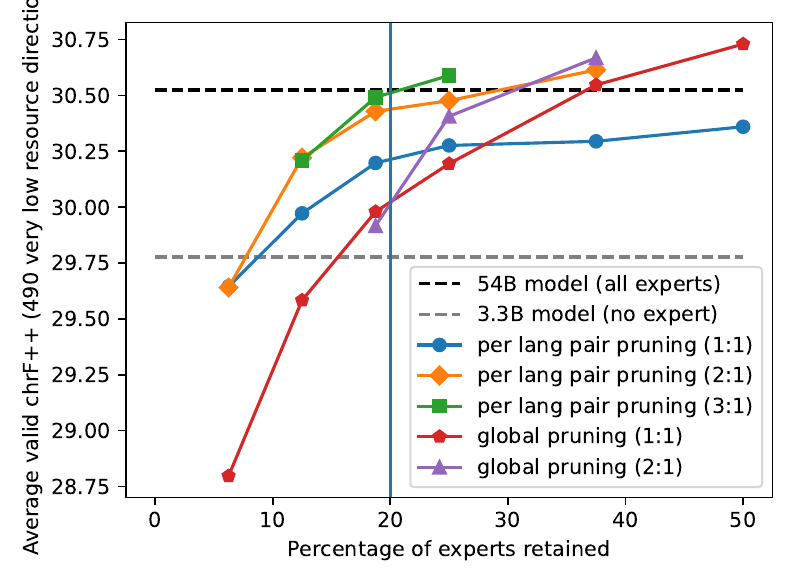}
    \caption{chrF++ and spBLEU valid scores on 30 languages for different resource types as a function of the percentage of experts retained. Pruning is done per language pair with the \textit{importance} metric and with a fixed number of experts per layer.}
    \label{fig:pruning_level}
\end{figure*}

\begin{table*}[t]
\begin{center}
\resizebox{\textwidth}{!}{
\begin{tabular}{ c | c | c | c | c | c | c | c | c | c | c | c | c  } 
Method & Enc experts & Dec experts & High$\rightarrow$High &High$\rightarrow$Low & High$\rightarrow$V. Low & Low$\rightarrow$High & Low$\rightarrow$Low & Low$\rightarrow$V. low & V. low$\rightarrow$High & V. low$\rightarrow$Low & V. low$\rightarrow$V. low & Average \\
\hline
3.3B dense model & 6 & 6 & 44.18 & 38.30 & 31.45 & 38.24 & 34.60 & 27.93 & 35.93 & 32.02 & 26.47 & 35.81 \\
54.5B MoE model & 768 & 768 & \textbf{45.41} & 38.98 & \textbf{31.89} & \textbf{39.72} & \textbf{35.40} & \textbf{28.83} & \textbf{37.29} & \textbf{33.23} & \textbf{26.95} & \textbf{36.81} \\
\hline
Fixed per layer (lang-pair) & 216 & 72 & 45.37 & 39.06 & 31.79 & 39.20 & 35.03 & 28.47 & 37.05 & 33.16 & 26.63 & 36.59 \\
Fixed per layer (global) & 216 & 72 & 43.20 & 37.60 & 31.68 & 37.37 & 33.94 & 28.40 & 35.38 & 31.97 & 26.84 & 35.34 \\
Fixed per layer (lang) & 216 & 72 & 45.35 & \textbf{39.10} & 31.82 & 39.18 & 35.10 & 28.51 & 37.02 & 33.19 & 26.62 & 36.61 \\

\end{tabular} 
}
\caption{chrF++ test scores on 53 languages, with the \textit{importance} metric for 80\% pruning (1-GPU decoding).}
\label{tab:test_results}
\end{center}
\end{table*}



\begin{table*}[t]
\begin{center}
\resizebox{\textwidth}{!}{
\begin{tabular}{ c | c | c | c | c | c | c | c | c | c | c | c | c  } 
Method & Enc experts & Dec experts & High$\rightarrow$High &High$\rightarrow$Low & High$\rightarrow$V. low & Low$\rightarrow$High & Low$\rightarrow$Low & Low$\rightarrow$V. low & V. low$\rightarrow$High & V. low$\rightarrow$Low & V. low$\rightarrow$V. low & Average \\
\hline
3.3B dense model & 6 & 6 & 45.54 & 38.84 & 32.72 & 39.18 & 34.87 & 29.07 & 38.39 & 34.11 & 29.21 & 34.64 \\
54.5B MoE model & 768 & 768 & \textbf{46.68} & 39.36 & \textbf{33.56} & \textbf{40.53} & 35.49 & \textbf{30.07} & \textbf{40.46} & \textbf{35.49} & \textbf{30.16} & \textbf{35.74} \\
Fixed per layer (lang) & 216 & 72 & 46.67 & \textbf{39.59} & 33.33 & 40.19 & \textbf{35.50} & 29.67 & 39.94 & 35.29 & 29.50 & 35.46 \\
\end{tabular} 
}
\caption{chrF++ test scores on all 202 languages, with the \textit{importance} metric for 80\% pruning (1-GPU decoding).}
\label{tab:full_test_results}
\end{center}
\end{table*}



\section{Experiments}
\label{sec: exp_res}


\subsection{Evaluation settings}

In our experiments, we use the FLORES-200 benchmark \cite{costa2022no}, which consists of translations of 3001 English sentences (from 842 distinct Wikipedia articles) to all other 201 languages. The multi-parallel nature of this dataset makes it possible to evaluate performance in all 40\,602 language directions.
As our final test benchmark, we take a representative subsample of 53 languages out of 202, which were also used as an ablation dataset by \citet{costa2022no}.
In our intermediate experiments, we work with a smaller subset of 30 out of 53 languages, with 10 languages per resource type (high, low, very low) and covering the same fourteen language families as the full subset of 53 languages. More details on the languages considered in our experiments as well as the amount of resources available per category are provided in Tables \ref{tab:all_langs} and \ref{tab:distrib} in Appendix.

\begin{table*}[!t]
\resizebox{\textwidth}{!}{
\begin{tabular}{|r|c|c|c|c|c|c|c|c|c|c|c|c|}
\hline
\diagbox{\cellcolor{blue!10}Encoder}{Decoder} & En$\rightarrow$Fr & En$\rightarrow$Ur & Ast$\rightarrow$Ur & Ur$\rightarrow$Fr & Ur$\rightarrow$Ast & Fr$\rightarrow$Ast & Ast$\rightarrow$Ko & Fr$\rightarrow$Ko& Ko$\rightarrow$En & Ko$\rightarrow$Ast & Fr$\rightarrow$En & Ast$\rightarrow$En \\
\hline
En$\rightarrow$Fr & NA & 0.18 & 0.17 & \textbf{0.71} & 0.31 & 0.32 & 0.17 & 0.17 & 0.19 & 0.31 & 0.17 & 0.17 \\
\hline
En$\rightarrow$Ur & \cellcolor{blue!10}\textbf{1.00} & NA & \textbf{0.68} & 0.20 & 0.24 & 0.22 & 0.31 & 0.30 & 0.26 & 0.23 & 0.21 & 0.23 \\
\hline
Ast$\rightarrow$Ur & \cellcolor{blue!10}0.35 & \cellcolor{blue!10}0.35 & NA & 0.22 & 0.25 & 0.22 & 0.37 & 0.35 & 0.21 & 0.22 & 0.18 & 0.20 \\
\hline
Ur$\rightarrow$Fr & \cellcolor{blue!10}0.36 & \cellcolor{blue!10}0.36 & \cellcolor{blue!10}0.40 & NA & 0.39 & 0.38 & 0.19 & 0.19 & 0.22 & 0.39 & 0.16 & 0.19 \\
\hline
Ur$\rightarrow$Ast & \cellcolor{blue!10}0.36 & \cellcolor{blue!10}0.36 & \cellcolor{blue!10}0.40 & \cellcolor{blue!10}\textbf{1.00} & NA & \textbf{0.83} & 0.27 & 0.25 & 0.24 & \textbf{0.87} & 0.17 & 0.22 \\
\hline
Fr$\rightarrow$Ast & \cellcolor{blue!10}0.42 & \cellcolor{blue!10}0.42 & \cellcolor{blue!10}0.45 & \cellcolor{blue!10}0.44 & \cellcolor{blue!10}0.44 & NA & 0.25 & 0.24 & 0.24 & \textbf{0.80} & 0.20 & 0.22 \\
\hline
Ast$\rightarrow$Ko & \cellcolor{blue!10}0.35 & \cellcolor{blue!10}0.35 & \cellcolor{blue!10}\textbf{1.00} & \cellcolor{blue!10}0.40 & \cellcolor{blue!10}0.40 & \cellcolor{blue!10}0.45 & NA & \textbf{0.78} & 0.15 & 0.24 & 0.15 & 0.16 \\
\hline
Fr$\rightarrow$Ko & \cellcolor{blue!10}0.42 & \cellcolor{blue!10}0.42 & \cellcolor{blue!10}0.45 & \cellcolor{blue!10}0.44 & \cellcolor{blue!10}0.44 & \cellcolor{blue!10}\textbf{1.00} & \cellcolor{blue!10}0.45 & NA & 0.14 & 0.23 & 0.15 & 0.13 \\
\hline
Ko$\rightarrow$En & \cellcolor{blue!10}0.33 & \cellcolor{blue!10}0.33 & \cellcolor{blue!10}0.41 & \cellcolor{blue!10}0.48 & \cellcolor{blue!10}0.48 & \cellcolor{blue!10}0.44 & \cellcolor{blue!10}0.41 & \cellcolor{blue!10}0.44 & NA & 0.26 & \textbf{0.55} & \textbf{0.70} \\
\hline
Ko$\rightarrow$Ast & \cellcolor{blue!10}0.33 & \cellcolor{blue!10}0.33 & \cellcolor{blue!10}0.41 & \cellcolor{blue!10}0.48 & \cellcolor{blue!10}0.48 & \cellcolor{blue!10}0.44 & \cellcolor{blue!10}0.41 & \cellcolor{blue!10}0.44 & \cellcolor{blue!10}\textbf{1.00} & NA & 0.17 & 0.22 \\
\hline
Fr$\rightarrow$En & \cellcolor{blue!10}0.42 & \cellcolor{blue!10}0.42 & \cellcolor{blue!10}0.45 & \cellcolor{blue!10}0.44 & \cellcolor{blue!10}0.44 & \cellcolor{blue!10}\textbf{1.00} & \cellcolor{blue!10}0.45 & \cellcolor{blue!10}\textbf{1.00} & \cellcolor{blue!10}0.44 & \cellcolor{blue!10}0.44 & NA & \textbf{0.61} \\
\hline
Ast$\rightarrow$En & \cellcolor{blue!10}0.35 & \cellcolor{blue!10}0.35 & \cellcolor{blue!10}\textbf{1.00} & \cellcolor{blue!10}0.40 & \cellcolor{blue!10}0.40 & \cellcolor{blue!10}0.45 & \cellcolor{blue!10}\textbf{1.00} & \cellcolor{blue!10}0.45 & \cellcolor{blue!10}0.41 & \cellcolor{blue!10}0.41 & \cellcolor{blue!10}0.45 & NA \\
\hline
\end{tabular}
}
\caption{\label{tab:sim_lang_pair} The Jaccard similarity of selected 25\% important experts between different language pairs in the encoder (lower triangle) and decoder (upper triangle). Pruning is done per language pair with the \textit{importance} metric. The same number of experts were chosen for the encoder and decoder with thresholding.  }
\end{table*}

To evaluate translation quality we use two metrics: chrF++\footnote{SacreBLEU signature for chrF++: \scriptsize{\tt nrefs:1|case:mixed\\|eff:yes|nc:6|nw:2|space:no|version:2.3.1}} \cite{popovic2015chrf} and spBLEU\footnote{SacreBLEU signature for spBLEU:  \scriptsize{\tt nrefs:1|case:mixed\\|eff:no|tok:flores200|smooth:exp|version:2.3.1}} \cite{costa2022no}. BLEU is heavily tokenization-dependant and its implementations do not include tokenizers for most of the NLLB-200 languages. spBLEU overcomes this issue by tokenizing the references and model outputs with a multilingual SentencePiece tokenizer (SPM-200, \citealp{costa2022no}).
We report chrF++ results in the main paper and spBLEU results in Appendix. We use FLORES-200 dev (which we call \emph{valid}) for collecting MoE gate statistics and comparing different pruning algorithms and rates, and FLORES-200 devtest (which we call \emph{test}) for reporting final results and comparing with the 3.3B and 54.5B baselines.

\subsection{Results}
\label{sec:results}

In the first set of experiments, we work with a subset of 30 languages. 
Table \ref{tab:pruning_algo_and_metric} compares different expert pruning metrics and strategies under a 75\% pruning rate. The experts are selected per language pair, and the scores are averaged per resource type (high, low, very low). The first part of the table reports two baselines: an upper bound corresponding to the full (unpruned) 54.5B MoE model, and a lower bound being the 3.3B dense model (same architecture without experts).

\paragraph{Pruning metric}

The second part of Table~\ref{tab:pruning_algo_and_metric} compares the chrF++ performance of different pruning metrics (spBLEU score are reported in Appendix Table \ref{tab:pruning_algo_and_metric_spbleu}). From these results, we can see that the \textit{top-1 activity} and \textit{importance} metrics are the most effective at identifying important experts.  Further experiments with global threshold pruning (third part of Table~\ref{tab:pruning_algo_and_metric}) confirm the slightly better performance of the \textit{importance} metric which we keep as the default for the next experiments. 

\paragraph{Pruning algorithm}
Table \ref{tab:pruning_algo_and_metric} also compares the pruning algorithms described in Section~\ref{sec:mechanism} (\textit{fixed per layer} and \textit{global threshold}). Note that with \textit{fixed per layer}, we can either allocate the same expert budget in the encoder and decoder (balanced setting) or have more experts in the encoder (unbalanced setting). 

First, we see that the \textit{global threshold} strategy gives the best results overall, with the same average chrF++ as the full unpruned model. However, \textit{global threshold} is not very practical for several reasons. First, it identifies a different amount of experts per layer for each language pair, which leads to variable memory usage across language pairs. It also requires recreating and reloading the model when decoding multiple directions, which is very slow. Finally, we found that it was more sensitive to over-generation and hallucinations (which we elaborate on in Section~\ref{sec:hallucinations} in Appendix) at higher pruning rates. The \textit{enc/dec thresholds} approach does not suffer from all the limitations of \textit{global threshold}, but it is not better than \textit{fixed per layer} either. Therefore, for simplicity, we pick the \textit{fixed per layer} approach for our next experiments.

\paragraph{Balanced versus unbalanced pruning}
When retaining 25\% of experts (384 out of 12$\times$128), \textit{global threshold} keeps on average 335 encoder experts and 49 decoder experts. The number of selected experts in the encoder and decoder for different language resource types is shown in Table \ref{tab:num_experts_encdec} in Appendix.
Following this observation that encoder experts seem more important than decoder ones, we experiment with different encoder/decoder ratios. \textbf{1:1} is the balanced setting. \textbf{2:1} and \textbf{3:1} are unbalanced with respectively twice and three times as many encoder experts as decoder experts. Figure~\ref{fig:pruning_level} shows that \textbf{3:1} performs the best across almost all pruning rates and resource types.

\paragraph{Pruning with global statistics.}
Figure~\ref{fig:pruning_level} and Figure~\ref{fig:pruning_level_spbleu} in Appendix also show that the same experts can be pruned across all language pairs (with statistics aggregated over all directions) with no loss in performance at 50\% pruning. Statistics at the language-direction granularity let us safely prune up to 80\% of the experts (in the unbalanced setting), which makes the model small enough to fit on a single GPU.

\paragraph{Test results and language-specific pruning.}
Finally, we validate our results over the test set on 53 languages (2\,756 directions). We use the \textit{fixed per layer} approach with a \textbf{3:1} ratio, which showed the best results on the validation set at 80\% (minimum rate for 1-GPU decoding). Tables~\ref{tab:test_results} and~\ref{tab:test_results_spbleu} report these test scores with three different levels of granularity: \textit{global}, \textit{language-pair-specific} or \textit{language-specific} (as described in Section~\ref{sec:granularity}). Table~\ref{tab:test_results_valid} in the Appendix reports valid scores with the same settings.

Pruning important experts chosen per language pair gives 0.8 chrF++ more on average than the 3.3B dense model, and 0.2 chrF++ less than the full MoE model. Global pruning on the other hand performs worse than both the MoE and dense models, which confirms the importance of having a language-specific pruning strategy.

While choosing important experts for each language pair is effective, it is not very practical: with $L$ languages, this generates $L\times(L-1)$ different configurations. A more practical approach is to prune encoder experts per source language and decoder experts per target language (i.e., \emph{language-specific} pruning). This pruning strategy performs exactly as well as pruning per language direction and is more convenient.
Following this observation, we extract per-language gate statistics on all 202 languages.\footnote{By decoding 25 random line pairs per language direction, resulting in 5\,025 lines per source language and per target language. To speed up this process, we do teacher forcing instead of beam-search decoding, which we found to perform as well.} Then, we apply 80\% per-layer pruning with the \textit{importance} metric (at the language granularity) and decode the test set in all 40\,602 directions. Tables~\ref{tab:full_test_results}~and~\ref{tab:full_test_results_spbleu} report the chrF++ and spBLEU scores. Table~\ref{tab:full_test_results_deltas} reports average score deltas with the unpruned model (and standard deviation per resource type). To facilitate future research and give the opportunity for anyone with a 32GB GPU to run the NLLB-200 model, we release the detailed gate statistics and the ids of the selected experts. We also share the scores for each direction and the decoding outputs of our best pruning approaches.


\section{Discussion}

\subsection{Inference speed and compute budget}
Table \ref{tab:inference} reports the inference speed of different models: the 3.3B dense model, the full MoE model, and the MoE model with 80\% pruning. We see that with 80\% pruning, the MoE model requires a single 32GB V100 and performs approximately as fast as the full model on 4 GPUs. If 4 GPUs are available, 80\% pruning can double the inference speed of the MoE model.\\
Table~\ref{tab:compute_power} in Appendix gives a breakdown of the number of GPU hours used for this work.

\begin{table}[t]
\centering
\small
\begin{tabular}{c|c|c|c|c}
Model & Batch size & GPUs & WPS & Time (s) \\
\hline
\multirow{2}{*}{54.5B} & \multirow{2}{*}{16k} & 8 & 195 & 105 \\
 & & 4 & 156 & 131 \\
\hline
\multirow{2}{*}{80\% pruning} & 16k & 4 & 299 & 79 \\
& 4k & 1 & 172 & 135 \\
\hline
3.3B & 4k & 1 & 246 & 86 \\
\end{tabular}
\caption{Inference speed benchmark for the 3.3B dense baseline model, the full MoE model, and its pruned version with 36 experts per encoder layer and 12 per decoder layer. We decode the FLORES valid set from 29 languages into English and average the decoding time or words per second.}
\label{tab:inference}
\end{table}

\subsection{Similarity of selected experts}
Section~\ref{sec:results} shows that only a fraction of all experts is necessary to translate between two given languages. 
We analyze the experts selected by our pruning method, to verify whether we can claim that there are indeed language-specific experts. 
In order to do so, we select experts with our proposed \textit{importance} metric and prune them per language pair at a 75\% rate with the \textit{Enc/dec thresholds} method, so that both the encoder and decoder have the same number of experts. We then compute the Jaccard similarity of selected encoder/decoder experts between different language pairs sharing the same source or target language. The lower and upper triangles of Table~\ref{tab:sim_lang_pair} show this similarity in the encoder and decoder respectively. We see that the encoder experts are independent of the target language (even though pruning is based on statistics collected at the lang-pair granularity level). This is an expected result, and it is due to the model design, where the target language code is introduced on the decoder side only: the encoder representation is not impacted by the target language. We note that the similarity between different source languages is also quite high (30-50\%). 
The similarity between important decoder experts for the same target language is in the 68-87\% range; and in the 13-39\% range for different target languages. These observations combined with the results in Section~\ref{sec:results} suggest the emergence of language-specific experts in the NLLB-200 model.

\subsection{Similarity of languages based on the importance metric}
\begin{figure}
    \centering
    \includegraphics[width=0.9\columnwidth]{"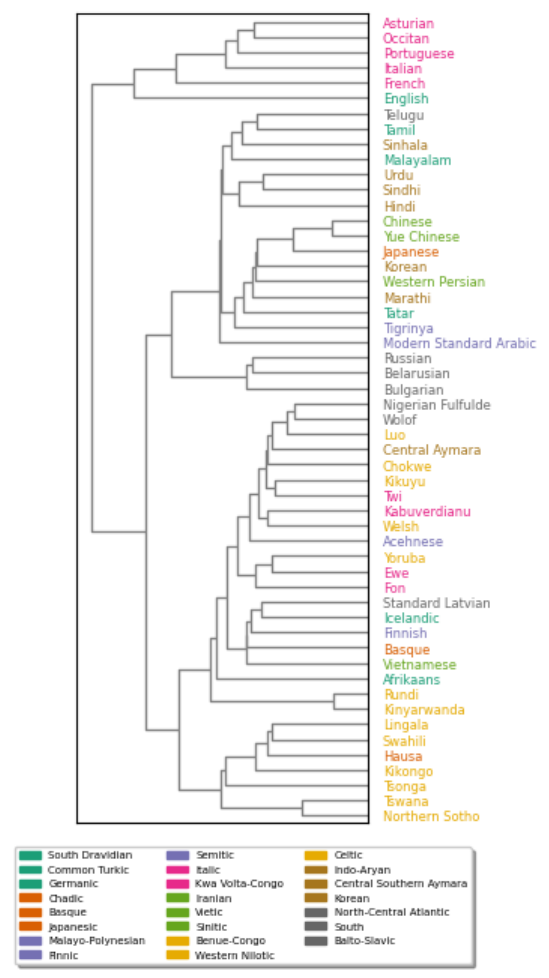"}
    \caption{Hierarchical clustering of languages based on the \textit{importance} metric of experts in the decoder. Different colors represent different language subgroupings.}
    \label{fig:language_dendrogram}
\end{figure}

Finally, we compare expert statistics across different languages, to better understand whether knowledge transfer happens at the expert level between similar languages. 
We gather importance metrics for each expert in the decoder for each language and concatenate the values of all MoE layers to have one feature vector of dimension 768. Then we do hierarchical clustering and show it as a dendrogram in Figure~\ref{fig:language_dendrogram}, where we highlight different language subgroupings with different colors. We can see that some clusters contain linguistically related languages, such as Yue Chinese, Korean and Japanese; Russian and Belarussian; or Portuguese, Asturian, and French. We run a similar analysis on the encoder experts and also observe meaningful language clustering, but less clear (Appendix Figure~\ref{fig:language_dendrogram_encoder}).

\subsection{Discrepancy between chrF++ and spBLEU scores} 
We observed that our pruning method results in slightly higher performance drop according to spBLEU, than with chrF++.
We hypothesize that it is due to a rare but visible phenomenon of over-generation (and sometimes hallucinations).  In the majority of cases, the translation is accurate initially but subsequently includes repetitions, paraphrasing, or slight hallucinations. The spBLEU metric penalizes this behavior more than chrF++, which could account for the variation in scores observed. More details on this are in Section~\ref{sec:hallucinations} in Appendix.

\section{Conclusion}
In this paper, we study expert pruning in the NLLB-200 Mixture-of-Experts MT model. We propose expert pruning metrics based on gate statistics collected while decoding. We study several pruning strategies and demonstrate that it is possible to prune up to 80\% of experts with a negligible loss in performance, which makes it possible to decode on a single 32GB GPU. We compare pruning at three levels of granularity: per language direction, per language, or global. Language-specific and language-pair pruning perform the same but the former is the most convenient. Global pruning (i.e., pruning always the same experts regardless of the source and target languages) performs surprisingly well but worse than language-specific pruning, which suggests that there are indeed some language-specific experts. This latter hypothesis is confirmed by our analysis of the selected experts.


\section{Risks and Limitations}
In our work, we rely on a single Mixture-of-Experts NMT model which is NLLB-200. There is a risk that our conclusions may only hold for this particular model and are specific to the way this model was trained. We believe that our findings still can be of interest to any person willing to use the NLLB-200 model because: 
    (1) It was the only publicly-available MoE NMT model at the time of submission;
    (2) It is the only model covering 202 languages and reaching SoTA results for most of those languages.
    
Moreover, we did not try to finetune the pruned model, which could potentially improve the results (but requires a large number of GPUs) and therefore change some of our conclusions. 

This work has similar risks as the original NLLB-200 models regarding the misuse of potentially wrong translations. Note that, as observed by \citet{Mohammadshahi2022WhatDC}, pruning could amplify the biases already present in the full model.

\section*{Acknowledgement}
This work was completed during a research internship at NAVER LABS Europe. Yeskendir Koishekenov is also supported by ELLIS Amsterdam and Qualcomm AI Research.

\bibliography{anthology,custom}

\appendix
\newpage 
\onecolumn
\section{Discrepancy between chrF++ and spBLEU scores}
\label{sec:hallucinations}

\begin{figure*}
    \centering
    \includegraphics[width=0.325\textwidth]{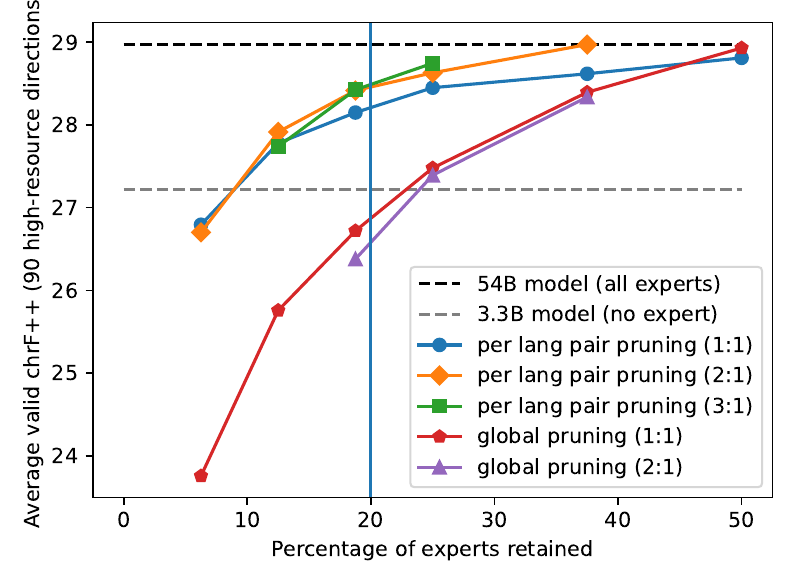}
    \includegraphics[width=0.325\textwidth]{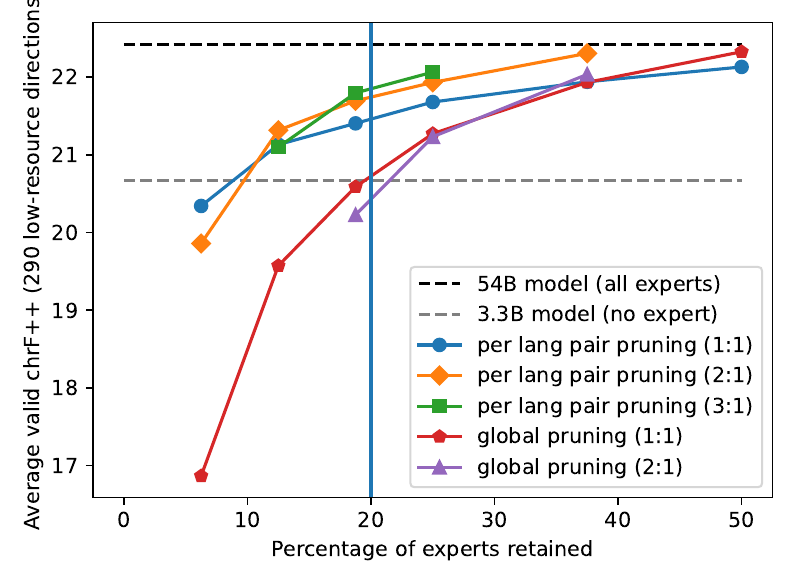}
    \includegraphics[width=0.325\textwidth]{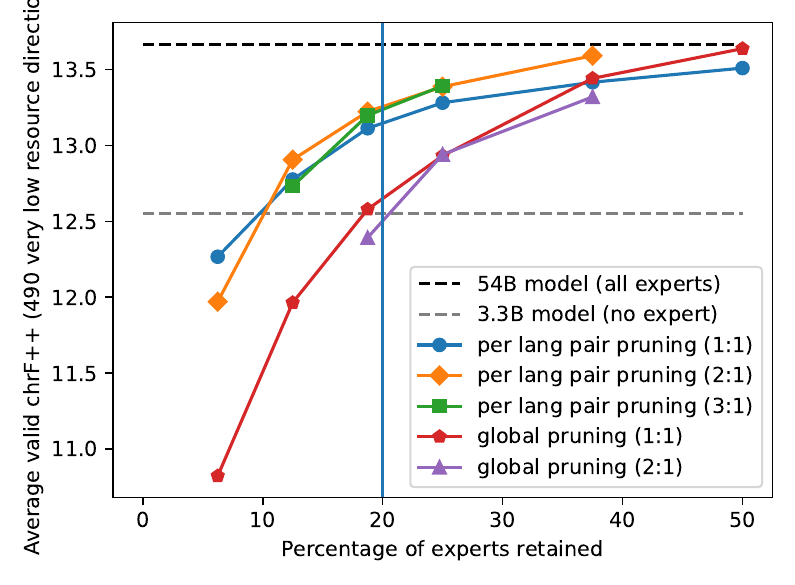}
    \caption{spBLEU valid scores on 30 languages for different resource types as a function of the percentage of experts retained. Pruning is done at the language pair granularity with the \textit{importance} metric and with a fixed number of experts per layer.}
    \label{fig:pruning_level_spbleu}
\end{figure*}

The spBLEU scores (Figure~\ref{fig:pruning_level} top right, or Figure~\ref{fig:pruning_level_spbleu} and Tables~\ref{tab:pruning_algo_and_metric_spbleu} and~\ref{tab:test_results_spbleu}) do not show exactly the same trend as chrF++. The gap between the full models and their pruned versions is slightly higher. This is likely caused by a rare but visible phenomenon of over-generation (and sometimes hallucinations). Table~\ref{tab:over_generation} shows some examples of such over-generation (with \textbf{3:1} \textit{fixed per layer} lang-pair pruning at 80\%). Most of the time, the translation is correct, but then continues with repetitions of itself, paraphrasing, or slight hallucinations. This behavior is more penalized by spBLEU than chrF++, which may explain the difference in scores. For instance, when duplicating the FLORES valid English-French translation output of the 54.5B model (i.e., concatenating each output sentence with itself), we see a spBLEU drop of 47\% and a chrF++ drop of only 13\%. The \emph{global threshold} method is more sensitive to this phenomenon. For instance, 80\% pruning leads to a 1.75 spBLEU drop (vs 0.53 for the \textit{fixed per layer} method). We 
report in Table~\ref{tab:len_ratios} the difference in length ratio (reported by SacreBLEU, \citealp{post-2018-call}) between the pruned models and the full model. We observe that \textit{global threshold} at 80\% has an average length ratio delta with the full model of 0.16 (meaning it generates longer outputs), while \emph{fixed per layer} has 0.04. We hypothesize that this over-generation issue may be mitigated by identifying experts that are specialized in generating the end-of-sequence symbol, but this is the subject of future work.

\begin{table*}[h]
\begin{center}
\resizebox{\textwidth}{!}{
\begin{tabular}{ c | c | c | c | c | c | c | c | c | c | c | c | c  } 
Method & Enc experts & Dec experts & High$\rightarrow$High &High$\rightarrow$Low & High$\rightarrow$V. Low & Low$\rightarrow$High & Low$\rightarrow$Low & Low$\rightarrow$V. low & V. low$\rightarrow$High & V. low$\rightarrow$Low & V. low$\rightarrow$V. low & Average \\
\hline
3.3B dense model & 6 & 6 & 0.02$\pm$0.02 & 0.04$\pm$0.03 & 0.09$\pm$0.06 & 0.02$\pm$0.03 & 0.04$\pm$0.04 & 0.11$\pm$0.06 & 0.06$\pm$0.05 & 0.07$\pm$0.07 & 0.15$\pm$0.08 & 0.06$\pm$0.07 \\
\hline
\multirow{3}{*}{Fixed per layer} & 288 & 96 & 0.03$\pm$0.03 & 0.01$\pm$0.02 & 0.04$\pm$0.03 & 0.04$\pm$0.06 & 0.01$\pm$0.03 & 0.04$\pm$0.03 & 0.06$\pm$0.06 & 0.03$\pm$0.03 & 0.06$\pm$0.04 & 0.04$\pm$0.04 \\
& 216 & 72 & 0.04$\pm$0.03 & 0.02$\pm$0.02 & 0.05$\pm$0.04 & 0.05$\pm$0.07 & 0.01$\pm$0.04 & 0.05$\pm$0.05 & 0.07$\pm$0.07 & 0.04$\pm$0.04 & 0.07$\pm$0.04 & 0.04$\pm$0.05 \\
& 144 & 48 & 0.05$\pm$0.05 & 0.03$\pm$0.04 & 0.07$\pm$0.05 & 0.07$\pm$0.09 & 0.05$\pm$0.10 & 0.08$\pm$0.06 & 0.09$\pm$0.07 & 0.07$\pm$0.09 & 0.10$\pm$0.06 & 0.07$\pm$0.07 \\
\hline
\multirow{3}{*}{Global threshold} & \multicolumn{2}{c|}{384} & 0.07$\pm$0.07 & 0.07$\pm$0.10 & 0.10$\pm$0.07 & 0.13$\pm$0.11 & 0.11$\pm$0.17 & 0.14$\pm$0.09 & 0.13$\pm$0.12 & 0.11$\pm$0.14 & 0.14$\pm$0.09 & 0.11$\pm$0.11 \\
& \multicolumn{2}{c|}{288} & 0.10$\pm$0.10 & 0.12$\pm$0.17 & 0.15$\pm$0.20 & 0.19$\pm$0.22 & 0.15$\pm$0.21 & 0.18$\pm$0.24 & 0.20$\pm$0.25 & 0.13$\pm$0.15 & 0.19$\pm$0.22 & 0.16$\pm$0.20 \\
& \multicolumn{2}{c|}{192} & 0.10$\pm$0.10 & 0.12$\pm$0.15 & 0.12$\pm$0.09 & 0.17$\pm$0.15 & 0.17$\pm$0.23 & 0.16$\pm$0.12 & 0.16$\pm$0.14 & 0.17$\pm$0.19 & 0.15$\pm$0.11 & 0.15$\pm$0.15 \\
\hline
Enc/dec thresholds & 216 & 72 & 0.05$\pm$0.04 & 0.03$\pm$0.03 & 0.06$\pm$0.04 & 0.08$\pm$0.09 & 0.03$\pm$0.04 & 0.06$\pm$0.05 & 0.09$\pm$0.07 & 0.06$\pm$0.06 & 0.08$\pm$0.04 & 0.06$\pm$0.06 \\
\hline
\end{tabular} 
}
\caption{Valid length ratio statistics on 30 languages. We compute the length ratio deltas w.r.t. the full unpruned model and report the mean and standard deviation in each language category. Values smaller than zero mean that the model generates longer sequences on average. We prune with the \textit{importance} metric, using statistics at the language pair granularity. Length ratios are obtained with SacreBLEU and the `flores200' tokenization.}
\label{tab:len_ratios}
\end{center}
\end{table*}

\begin{table*}[h]
\begin{center}
\begin{tabular}{ c | p{13cm} }
Reference & However, very few Royal Navy ships were based near the likely invasion routes as the admirals were afraid they would be sunk by German air attack. \\
Translation & However, the Royal Navy had few ships in the vicinity of the invasion routes, as the admirals feared that they would be sunk by German air attack. \textcolor{red}{The Royal Navy's ships were too small to be able to operate in the vicinity of the invasion routes.} \\
\hline
Reference & However, the discovery of his tomb in 1922 made him a celebrity. While many tombs of the past were robbed, this tomb was left virtually undisturbed. \\
Translation & However, his tomb became famous when it was discovered in 1922. Although many tombs in the past have been excavated, this one has remained largely undisturbed. \textcolor{red}{The tomb was built in the middle of the 19th century. The tomb was built in the middle of the 20th century. The tomb was built in the middle of the 20th century} \\
\hline
Reference & The translation engines have improved dramatically, and now often give more or less correct translations (and more seldom gibberish), but some care is due, as they still may have gotten it all wrong. \\
Translation & Translation engines have improved dramatically, and now often produce more or less accurate translations (as well as more or less uncommon ones), but some errors can still be corrected, so some caution is warranted. \textcolor{red}{For example, in the case of the English language, the use of the word "translate" in the context of a translation of a document, such as a translation of a document into a foreign language, is not recommended. The use of the word "translate" in the context of a translation of a document is discouraged.} \\
\end{tabular} 
\caption{Examples of over-generation when translating from Tigrinya (\texttt{tir\_Ethi}) to English (\texttt{eng\_Latn}) with \textit{fixed per layer} (\textbf{3:1}) 80\% pruning at the language direction granularity.}
\label{tab:over_generation}
\end{center}
\end{table*}

\begin{table*}
\centering
\resizebox{0.9\linewidth}{!}{
\begin{tabular}{l| l| l| l| l}
Code & Language & Resource Type & Family & Subgrouping \\
\hline
tsn\_Latn & Tswana & High & Atlantic-Congo & Benue-Congo \\
vie\_Latn & Vietnamese & High & Austroasiatic & Vietic \\
rus\_Cyrl & Russian & High & Indo-European & Balto-Slavic \\
eng\_Latn & English & High & Indo-European & Germanic \\
fra\_Latn & French & High & Indo-European & Italic \\
por\_Latn & Portuguese & High & Indo-European & Italic \\
jpn\_Jpan & Japanese & High & Japonic & Japanesic \\
kor\_Hang & Korean & High & Koreanic & Korean \\
fin\_Latn & Finnish & High & Uralic & Finnic \\
tir\_Ethi & Tigrinya & Low & Afro-Asiatic & Semitic \\
nso\_Latn & Northern Sotho & Low & Atlantic-Congo & Benue-Congo \\
yor\_Latn & Yoruba & Low & Atlantic-Congo & Benue-Congo \\
mal\_Mlym & Malayalam & Low & Dravidian & South Dravidian \\
tam\_Taml & Tamil & Low & Dravidian & South Dravidian \\
bel\_Cyrl & Belarusian & Low & Indo-European & Balto-Slavic \\
cym\_Latn & Welsh & Low & Indo-European & Celtic \\
urd\_Arab & Urdu & Low & Indo-European & Indo-Aryan \\
luo\_Latn & Luo & Low & Nilotic & Western Nilotic \\
tat\_Cyrl & Tatar & Low & Turkic & Common Turkic \\
cjk\_Latn & Chokwe & Very low & Atlantic-Congo & Benue-Congo \\
kik\_Latn & Kikuyu & Very low & Atlantic-Congo & Benue-Congo \\
fuv\_Latn & Nigerian Fulfulde & Very low & Atlantic-Congo & North-Central Atlantic \\
wol\_Latn & Wolof & Very low & Atlantic-Congo & North-Central Atlantic \\
ace\_Latn & Acehnese & Very low & Austronesian & Malayo-Polynesian \\
ayr\_Latn & Central Aymara & Very low & Aymaran & Central Southern Aymara \\
snd\_Arab & Sindhi & Very low & Indo-European & Indo-Aryan \\
ast\_Latn & Asturian & Very low & Indo-European & Italic \\
kea\_Latn & Kabuverdianu & Very low & Indo-European & Italic \\
yue\_Hant & Yue Chinese & Very low & Sino-Tibetan & Sinitic \\
\hline
arb\_Arab & Modern Standard Arabic & High & Afro-Asiatic & Semitic \\
swh\_Latn & Swahili & High & Atlantic-Congo & Benue-Congo \\
eus\_Latn & Basque & High & Basque & Basque \\
bul\_Cyrl & Bulgarian & High & Indo-European & Balto-Slavic \\
lvs\_Latn & Standard Latvian & High & Indo-European & Balto-Slavic \\
afr\_Latn & Afrikaans & High & Indo-European & Germanic \\
isl\_Latn & Icelandic & High & Indo-European & Germanic \\
hin\_Deva & Hindi & High & Indo-European & Indo-Aryan \\
pes\_Arab & Western Persian & High & Indo-European & Iranian \\
ita\_Latn & Italian & High & Indo-European & Italic \\
zho\_Hans & Chinese & High & Sino-Tibetan & Sinitic \\
hau\_Latn & Hausa & Low & Afro-Asiatic & Chadic \\
kin\_Latn & Kinyarwanda & Low & Atlantic-Congo & Benue-Congo \\
kon\_Latn & Kikongo & Low & Atlantic-Congo & Benue-Congo \\
lin\_Latn & Lingala & Low & Atlantic-Congo & Benue-Congo \\
run\_Latn & Rundi & Low & Atlantic-Congo & Benue-Congo \\
tso\_Latn & Tsonga & Low & Atlantic-Congo & Benue-Congo \\
ewe\_Latn & Ewe & Low & Atlantic-Congo & Kwa Volta-Congo \\
fon\_Latn & Fon & Low & Atlantic-Congo & Kwa Volta-Congo \\
twi\_Latn & Twi & Low & Atlantic-Congo & Kwa Volta-Congo \\
tel\_Telu & Telugu & Low & Dravidian & South \\
mar\_Deva & Marathi & Low & Indo-European & Indo-Aryan \\
sin\_Sinh & Sinhala & Low & Indo-European & Indo-Aryan \\
oci\_Latn & Occitan & Very low & Indo-European & Italic \\

\end{tabular}
}
\caption{Set of 53 languages used in the experiments. We show the lang code, name, resource type, family, and subgrouping of each language. The 30 languages used in the intermediate experiments are in the top half of the table.}
\label{tab:all_langs}
\end{table*}


\begin{table*}
\begin{center}
\resizebox{\textwidth}{!}{
\begin{tabular}{ c | c | c | c | c | c | c | c | c | c | c | c  } 
Method & Metric & High$\rightarrow$High &High$\rightarrow$Low & High$\rightarrow$V. Low & Low$\rightarrow$High & Low$\rightarrow$Low & Low$\rightarrow$V. low & V. low$\rightarrow$High & V. low$\rightarrow$Low & V. low$\rightarrow$V. low & Average \\
\hline
\multicolumn{2}{c|}{3.3B dense model \cite{costa2022no}} & 27.22 & 20.77 & 11.29 & 23.10 & 17.85 & 9.44 & 19.07 & 14.65 & 7.84 & 16.78 \\
\multicolumn{2}{c|}{54.5B MoE model \cite{costa2022no}} & 28.98 & 22.29 & 11.87 & 25.19 & 19.49 & 10.24 & 20.79 & 16.55 & 8.36 & \textbf{18.17} \\
\hline
\multirow{5}{*}{Fixed per layer (balanced)} & Top 1 & 28.39 & 21.82 & 11.64 & 24.22 & 18.67 & 9.92 & 20.05 & 15.98 & 8.13 & 17.62 \\
& Top 2 & 27.06 & 20.89 & 11.20 & 23.08 & 17.87 & 9.48 & 19.08 & 15.18 & 7.83 & 16.82 \\
& Load balancing & 27.16 & 21.04 & 11.30 & 23.17 & 17.98 & 9.60 & 19.14 & 15.24 & 7.88 & 16.92 \\
& Importance (vanilla) & 25.92 & 18.27 & 10.51 & 23.78 & 17.64 & 9.69 & 19.20 & 14.43 & 7.73 & 16.33 \\
& Importance & 28.45 & 21.86 & 11.66 & 24.25 & 18.62 & 9.90 & 20.13 & 16.02 & 8.19 & 17.65 \\
\hline
\multirow{2}{*}{Global threshold} & Top 1 & 28.33 & 21.50 & 11.26 & 23.54 & 18.16 & 9.26 & 19.72 & 15.45 & 7.69 & 17.18 \\
& Importance & 28.43 & 21.56 & 11.28 & 23.52 & 18.37 & 9.40 & 19.74 & 15.54 & 7.69 & 17.25 \\
\hline
Fixed per layer (unbalanced) & \multirow{3}{*}{Importance} & 28.63 & 22.08 & 11.76 & 24.47 & 18.94 & 10.03 & 20.19 & 16.21 & 8.23 & 17.81 \\
Enc/Dec thresholds (balanced) & & 28.47 & 21.87 & 11.65 & 24.25 & 18.61 & 9.90 & 20.11 & 15.97 & 8.15 & 17.64 \\
Enc/Dec thresholds (unbalanced) & & 28.72 & 22.08 & 11.74 & 24.57 & 18.99 & 10.01 & 20.26 & 16.14 & 8.20 & 17.83 \\
\hline
\end{tabular}
}
\caption{spBLEU valid scores on 30 languages of different pruning algorithms and metrics, with 75\% pruning (i.e., 384 experts are kept in total). The unbalanced approaches keep 240 encoder experts and 144 decoder experts.}
\label{tab:pruning_algo_and_metric_spbleu}
\end{center}
\end{table*}

\begin{table*}
\begin{center}
\resizebox{\textwidth}{!}{
\begin{tabular}{ c | c | c | c | c | c | c | c | c | c | c | c | c  } 
Method & Enc experts & Dec experts & High$\rightarrow$High &High$\rightarrow$Low & High$\rightarrow$V. Low & Low$\rightarrow$High & Low$\rightarrow$Low & Low$\rightarrow$V. low & V. low$\rightarrow$High & V. low$\rightarrow$Low & V. low$\rightarrow$V. low & Average \\
\hline
3.3B dense model & 6 & 6 & 44.54 & 38.20 & 30.08 & 40.49 & 35.19 & 27.61 & 35.27 & 30.68 & 24.75 & 34.06 \\
54.5B MoE model & 768 & 768 & \textbf{45.90} & 39.19 & 30.24 & \textbf{42.29} & \textbf{36.35} & \textbf{28.18} & \textbf{36.55} & \textbf{32.16} & 24.93 & \textbf{35.07} \\
\hline
Fixed per layer (lang-pair) & 216 & 72 & 45.87 & 39.16 & 30.41 & 41.75 & 35.89 & 28.11 & 36.34 & 31.97 & 25.08 & 34.93 \\
Fixed per layer (global) & 216 & 72 & 44.56 & 37.80 & 29.91 & 40.61 & 34.78 & 27.85 & 35.51 & 31.04 & 25.09 & 34.10 \\
Fixed per layer (lang) & 216 & 72 & 45.84 & \textbf{39.22} & \textbf{30.46} & 41.72 & 35.96 & 28.17 & 36.29 & 32.03 & \textbf{25.11} & 34.96 \\
\hline
Enc/dec thresholds (lang-pair) & 216 & 72 & 45.89 & 39.19 & 30.39 & 41.77 & 36.02 & 28.21 & 36.28 & 31.97 & 25.07 & 34.95 \\
\cline{2-3}
Global threshold (lang-pair) & \multicolumn{2}{c|}{288} & 45.82 & 38.96 & 30.06 & 41.44 & 35.98 & 27.92 & 35.90 & 31.83 & 24.72 & 34.71 \\
\hline
\end{tabular} 
}
\caption{chrF++ valid scores on 30 languages, with the \textit{importance} metric for 80\% pruning (1-GPU decoding) at three different levels of granularity (global, per language or per language direction).}
\label{tab:test_results_valid}
\end{center}
\end{table*}

\begin{table*}
\begin{center}
\resizebox{\textwidth}{!}{
\begin{tabular}{ c | c | c | c | c | c | c | c | c | c | c | c | c  } 
Method & Enc experts & Dec experts & High$\rightarrow$High &High$\rightarrow$Low & High$\rightarrow$V. Low & Low$\rightarrow$High & Low$\rightarrow$Low & Low$\rightarrow$V. low & V. low$\rightarrow$High & V. low$\rightarrow$Low & V. low$\rightarrow$V. low & Average \\
\hline
3.3B dense model & 6 & 6 & 26.72 & 18.69 & 12.62 & 21.08 & 15.65 & 10.12 & 19.52 & 14.03 & 9.27 & 17.71 \\
54.5B MoE model & 768 & 768 & \textbf{28.42} & \textbf{20.11} & \textbf{13.31} & \textbf{22.81} & \textbf{16.93} & \textbf{10.99} & \textbf{21.35} & \textbf{15.75} & \textbf{9.91} & \textbf{19.12} \\
\hline
Fixed per layer (lang-pair) & 216 & 72 & 28.01 & 19.81 & 12.81 & 21.94 & 16.33 & 10.37 & 20.63 & 15.27 & 9.27 & 18.56 \\
Fixed per layer (global) & 216 & 72 & 24.15 & 17.15 & 12.26 & 18.34 & 13.89 & 9.77 & 17.41 & 12.78 & 8.88 & 16.03 \\
Fixed per layer (lang) & 216 & 72 & 27.87 & 19.82 & 12.78 & 21.79 & 16.37 & 10.35 & 20.51 & 15.28 & 9.19 & 18.50\\
\end{tabular} 
}
\caption{spBLEU test scores on 53 languages, with the \textit{importance} metric for 80\% pruning (1-GPU decoding) at three different levels of granularity (global, per language or per language direction).}
\label{tab:test_results_spbleu}
\end{center}
\end{table*}

\begin{table*}
\begin{center}
\resizebox{\textwidth}{!}{
\begin{tabular}{ c | c | c | c | c | c | c | c | c | c | c | c | c  } 
Method & Enc experts & Dec experts & High$\rightarrow$High &High$\rightarrow$Low & High$\rightarrow$V. Low & Low$\rightarrow$High & Low$\rightarrow$Low & Low$\rightarrow$V. low & V. low$\rightarrow$High & V. low$\rightarrow$Low & V. low$\rightarrow$V. low & Average \\
\hline
3.3B dense model & 6 & 6 & 26.86 & 18.35 & 14.18 & 20.91 & 15.15 & 11.48 & 20.09 & 14.38 & 11.42 & 15.95 \\
54.5B MoE model & 768 & 768 & \textbf{28.61} & \textbf{19.49} & \textbf{15.41} & \textbf{22.66} & \textbf{16.22} & \textbf{12.69} & \textbf{22.71} & \textbf{16.18} & \textbf{12.71} & \textbf{17.48} \\
Fixed per layer (lang) & 216 & 72 & 28.27 & 19.26 & 15.08 & 22.02 & 15.84 & 12.24 & 21.90 & 15.62 & 12.05 & 16.97 \\
\end{tabular} 
}
\caption{spBLEU test scores on all 202 languages, with the \textit{importance} metric for 80\% pruning (1-GPU decoding) at the language granularity.}
\label{tab:full_test_results_spbleu}
\end{center}
\end{table*}

\begin{table*}
\begin{center}
\resizebox{\textwidth}{!}{
\begin{tabular}{ c | c | c | c | c | c | c | c | c | c | c | c | c  } 
Method & Enc experts & Dec experts & High$\rightarrow$High &High$\rightarrow$Low & High$\rightarrow$V. Low & Low$\rightarrow$High & Low$\rightarrow$Low & Low$\rightarrow$V. low & V. low$\rightarrow$High & V. low$\rightarrow$Low & V. low$\rightarrow$V. low & Average \\
\hline
3.3B dense model & 6 & 6 & 1.14$\pm$1.23 & 0.52$\pm$1.15 & 0.84$\pm$2.17 & 1.34$\pm$1.30 & 0.62$\pm$1.23 & 1.00$\pm$2.24 & 2.07$\pm$1.46 & 1.38$\pm$1.62 & 0.95$\pm$2.23 & 1.10$\pm$1.83 \\
Fixed per layer (lang) & 216 & 72 & 0.01$\pm$0.58 & -0.23$\pm$0.55 & 0.23$\pm$1.55 & 0.34$\pm$0.58 & -0.01$\pm$0.60 & 0.39$\pm$1.61 & 0.53$\pm$0.60 & 0.19$\pm$0.57 & 0.66$\pm$1.68 & 0.29$\pm$1.17 \\
\hline
\hline
3.3B dense model & 6 & 6 & 1.75$\pm$1.35 & 1.13$\pm$1.25 & 1.24$\pm$2.07 & 1.76$\pm$1.42 & 1.07$\pm$1.21 & 1.21$\pm$1.89 & 2.62$\pm$1.67 & 1.80$\pm$1.58 & 1.29$\pm$1.95 & 1.53$\pm$1.74 \\
Fixed per layer (lang) & 216 & 72 & 0.33$\pm$0.92 & 0.22$\pm$0.56 & 0.34$\pm$1.25 & 0.65$\pm$1.08 & 0.38$\pm$0.53 & 0.44$\pm$1.18 & 0.81$\pm$1.00 & 0.56$\pm$0.52 & 0.66$\pm$1.16 & 0.51$\pm$0.99 \\
\end{tabular} 
}
\caption{Test chrF++ deltas (first part) and spBLEU deltas (second part) with the unpruned MoE model on all 202 languages. The pruned version uses the \textit{importance} metric with 80\% pruning at the language granularity. Each column reports the average score for a given language category, as well as the standard deviation. A positive value means that this model is worse than the full 54.5B model. The last column reports the average score and standard deviation over all 202$\times$201 directions.}
\label{tab:full_test_results_deltas}
\end{center}
\end{table*}

\newpage

\begin{figure*}[!htb]
    \centering
    \includegraphics[width=\linewidth]{"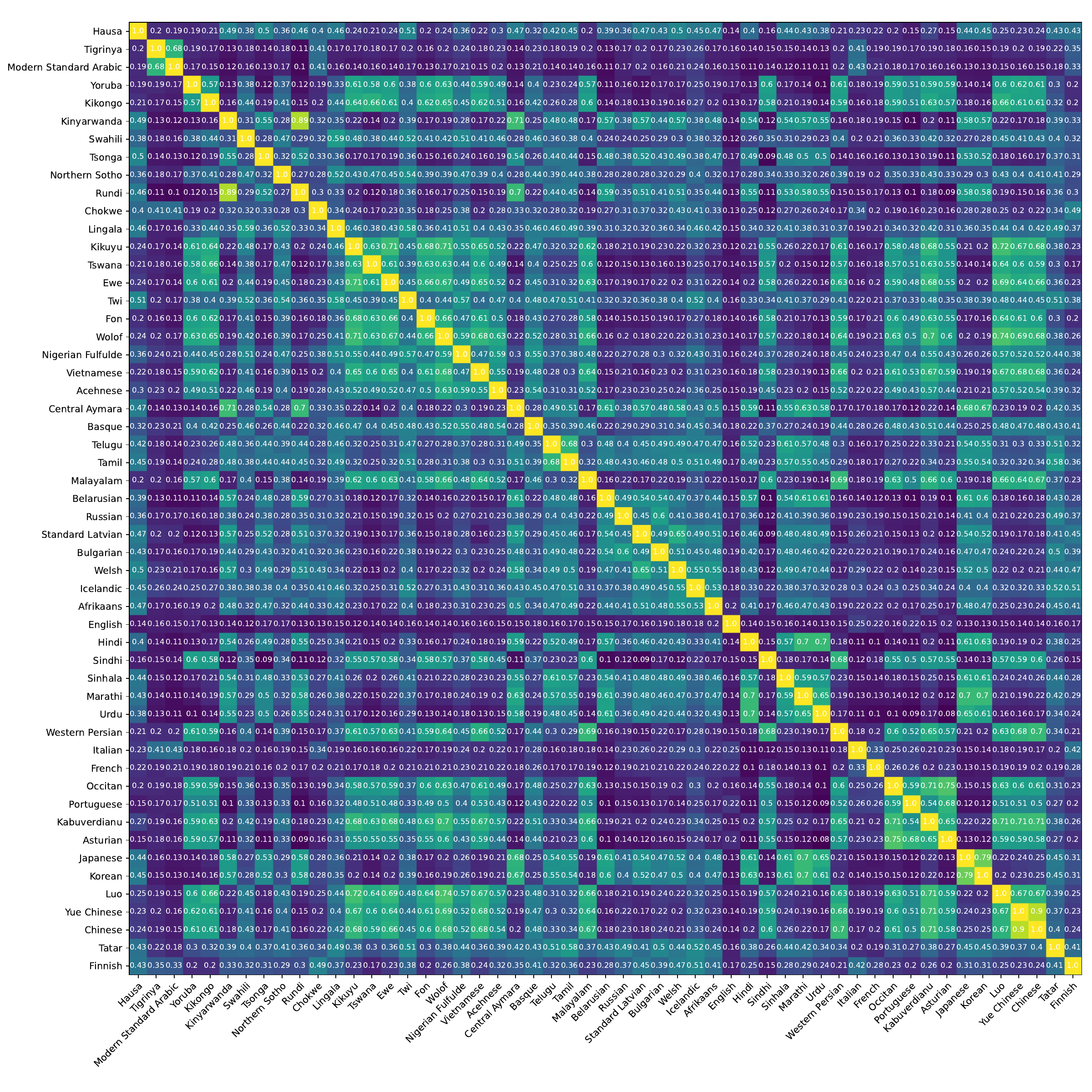"}
    \caption{Jaccard similarity of selected 25\% decoder experts for different languages. Pruning was done per language with the \textit{importance} metric and \textit{enc/dec threshold} pruning. Languages are sorted by language family.}
    \label{fig:language_similarity}
\end{figure*}

\begin{table}[!htb]
\centering
\begin{tabular}{ l c c  } 
Resource Type & Criterion & Language count \\
\hline
Very low & $|L| \leq 100k$ & 11 \\ 
Low & $ 100k \leq |L| \leq 1m$ & 22 \\ 
High & $ 1m \leq |L|$ & 20 \\ 
\hline
\end{tabular}
\caption{Distribution of languages in the 53-language subset, based on the amount of available data $|L|$. The 30-language subset has 10 languages of each resource type. Line counts are published by \citet{costa2022no} here: \url{https://tinyurl.com/535f7ust}}
\label{tab:distrib}
\end{table}

\newpage

\begin{figure*}[!htb]
    \centering
    \includegraphics[width=\linewidth]{"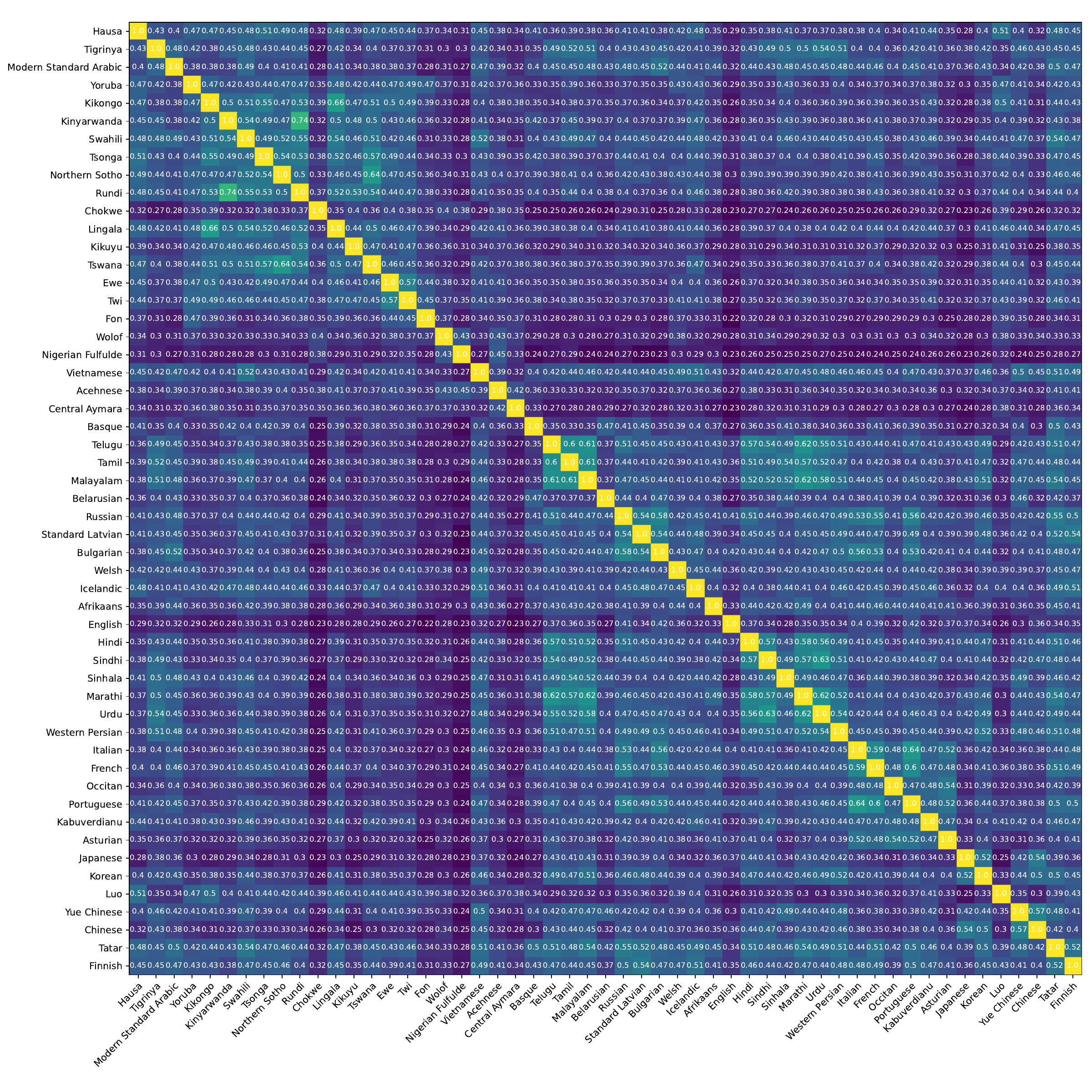"}
    \caption{Jaccard similarity of selected 25\% encoder experts for different languages. Pruning was done per language with the \textit{importance} metric and \textit{enc/Dec threshold} pruning. Languages are sorted by language family.}
    \label{fig:language_similarity_encoder}
\end{figure*}

\begin{table}[!htb]
\centering
\small
\begin{tabular}{c|c|c}
Model & Hours & GPU hours \\
\hline
3.3B & 480 & 440 \\
54.5B (full) & 4\,740 & 3\,840 \\
54.5B (pruned) & 15\,900 & 5\,700 \\
\hline
Total & 21\,120 & 9\,980 \\
\end{tabular}
\caption{Time spent decoding with each type of model in this work. This includes failed or non-discussed experiments. The ``hours'' column measures the total time spent by the decoding script, including model creation and loading (note that the GPUs were reserved but idle during that time). ``GPU hours'' measures the time actually spent decoding (i.e., with the GPU active).}
\label{tab:compute_power}
\end{table}

\newpage

\twocolumn

\begin{table}
\begin{center}
\resizebox{\columnwidth}{!}{
\begin{tabular}{ l | c | c  } 
Language pair resource type & Encoder & Decoder \\
\hline
High$\rightarrow$High & 320 & 64 \\
High$\rightarrow$Low & 344 & 44 \\
High$\rightarrow$V. low & 348 & 36 \\
\hline
Low$\rightarrow$High & 319 & 65 \\
Low$\rightarrow$Low & 343 & 41 \\
Low$\rightarrow$V. low & 346 & 38 \\
\hline
V. low$\rightarrow$High & 314 & 70 \\
V. low$\rightarrow$Low & 338 & 46 \\
V. low$\rightarrow$V. low & 340 & 44 \\
\hline
Average & 335 & 49 \\
\end{tabular} 
}
\caption{Average number of experts in the encoder and decoder for different language resource type language pairs with \textit{global threshold} 75\% pruning and the \textit{importance} metric.}
\label{tab:num_experts_encdec}
\end{center}
\end{table}

\begin{figure}[htp!]
    \centering
    \includegraphics[width=0.9\columnwidth]{"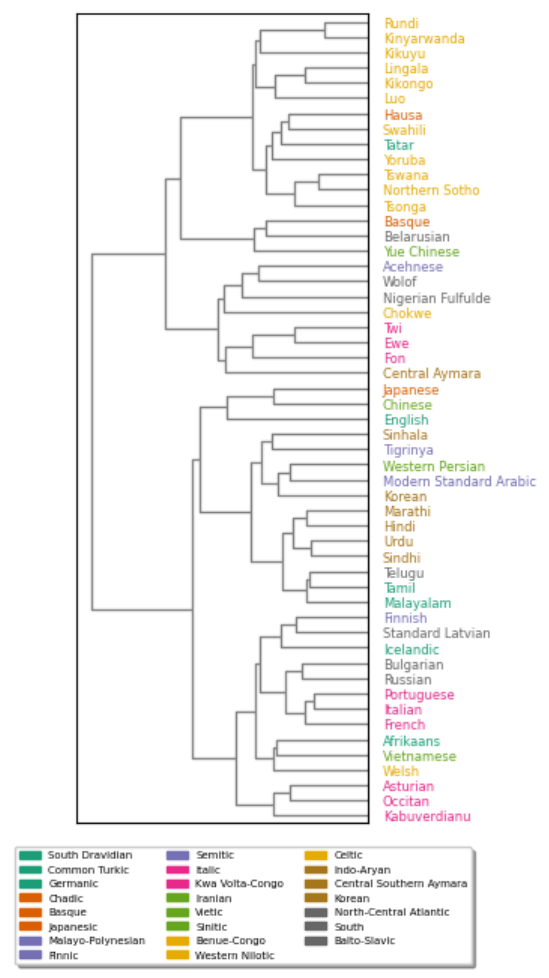"}
    \caption{Hierarchical clustering of languages based on the \textit{importance} metric of encoder experts. Different colors represent different language subgroupings.}
    \label{fig:language_dendrogram_encoder}
\end{figure}

\begin{figure}[htp!]
    \centering
    \includegraphics[width=0.7\columnwidth]{"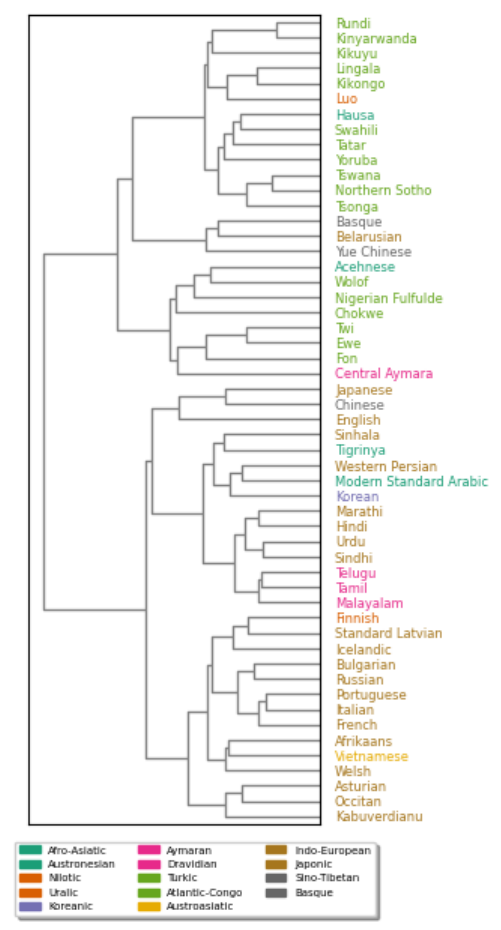"}
    \caption{Hierarchical clustering of languages based on the \textit{importance} metric of encoder experts. Different colors represent different language families.}
    \label{fig:language_dendrogram_encoder_family}
\end{figure}

\begin{figure}[htp!]
    \centering
    \includegraphics[width=0.7\columnwidth]{"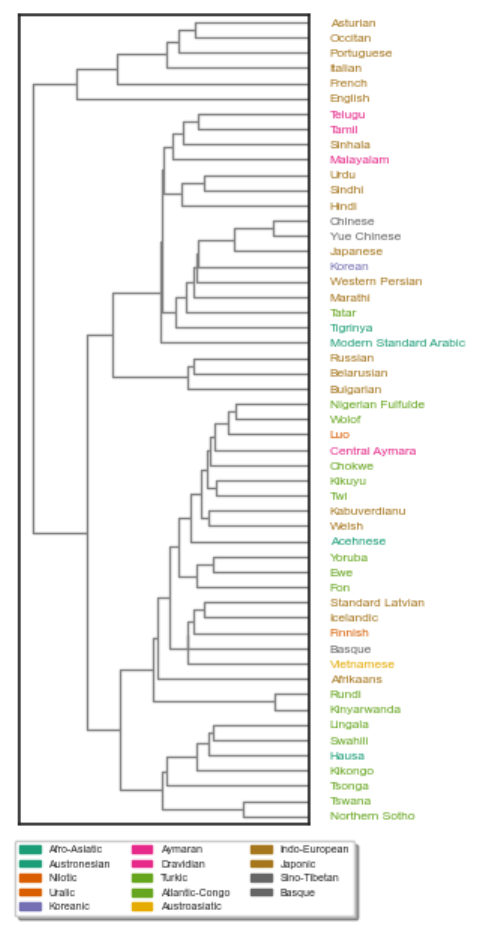"}
    \caption{Hierarchical clustering of languages based on the \textit{importance} metric of decoder experts. Different colors represent different language families.}
    \label{fig:language_dendrogram_decoder_family}
\end{figure}

\end{document}